\documentclass[letterpaper, 10 pt, conference]{ieeeconf}  
\IEEEoverridecommandlockouts
\usepackage[utf8]{inputenc}
\usepackage[english]{babel}
\usepackage[noend]{algpseudocode}
\usepackage{amsmath}
\usepackage{amssymb}
\usepackage{xcolor}
\usepackage{graphicx}
\usepackage{hyperref}
\usepackage{booktabs}
\usepackage{color}
\usepackage{array}
\usepackage{float}
\usepackage{stfloats}
\usepackage{multirow}
\usepackage{makecell}
\usepackage{threeparttable}
\usepackage[space]{cite}
\usepackage[linesnumbered,boxed,ruled,commentsnumbered]{algorithm2e}
\title{\LARGE \bf
Joint Intrinsic and Extrinsic LiDAR-Camera Calibration in 
Targetless \\ Environments Using Plane-Constrained Bundle Adjustment
}

\author{
Liang Li$^{1}$,
Haotian Li$^{1}$,
Xiyuan Liu$^{1}$,
Dongjiao He$^{1}$,
Ziliang Miao$^{1}$,
\\Fanze Kong$^{1}$,
Rundong Li$^{1}$, 
Zheng Liu$^{1}$,
and Fu Zhang$^{1*}$ 
\thanks{$^{*}$Corresponding author.}
\thanks{$^{1}$Liang Li, Haotian Li, Xiyuan Liu, Dongjiao He, Ziliang Miao, Fanze Kong, Rundong Li, Zheng Liu and Fu Zhang are with the Department of Mechanical Engineering, The University of Hong Kong, Hong Kong S.A.R., China. 
{\tt\small $\{$llihku, haotianl, xliuaa, hdj65822, miaozl, kongfz, rdli10010, u3007335$\}$@connect.hku.hk, fuzhang@hku.hk}.}
}

\begin{document}

\maketitle
\thispagestyle{empty}
\pagestyle{empty}

\begin{abstract}

This paper introduces a novel targetless method for joint intrinsic and extrinsic calibration of LiDAR-camera systems using plane-constrained bundle adjustment (BA). Our method leverages LiDAR point cloud measurements from planes in the scene, alongside visual points derived from those planes. The core novelty of our method lies in the integration of visual BA with the registration between visual points and LiDAR point cloud planes, which is formulated as a unified optimization problem. This formulation achieves concurrent intrinsic and extrinsic calibration, while also imparting depth constraints to the visual points to enhance the accuracy of intrinsic calibration. Experiments are conducted on both public data sequences and self-collected dataset. The results showcase that our approach not only surpasses other state-of-the-art (SOTA) methods but also maintains remarkable calibration accuracy even within challenging environments. For the benefits of the robotics community, we have open sourced our codes\footnote[2]{\url{https://github.com/hku-mars/joint-lidar-camera-calib}}.

\end{abstract}

\vspace{-1mm}
\section{INTRODUCTION}
\vspace{-1mm}

Autonomous robots frequently employ light detection and ranging (LiDAR) sensors and cameras for perception tasks, such as simultaneous localization and mapping (SLAM) \cite{mur2015orb,he2023point}, object detection \cite{redmon2016you,ye2020hvnet}, and semantic segmentation \cite{alonso2020mininet,milioto2019rangenet++}. As the direct depth measurements provided by LiDARs and the rich texture information captured by cameras are complementary to each other, LiDAR-camera fusion becomes a promising avenue to enhance the perception system performance. Therefore, research interest has surged in the development of robust and accurate LiDAR-camera fusion methods \cite{chou2021efficient,zheng2022fast,liu2022bevfusion,zhao2023lif} beyond vision-only or LiDAR-only approaches.

A crucial step in this fusion is estimating the 6 degrees-of-freedom (DoF) transformation—rotation and translation—between LiDAR and camera coordinate systems, which is known as LiDAR-camera extrinsic calibration. Among SOTA extrinsic calibration methods, target-based ones prove to be accurate and robust. These methods employ specially designed calibration targets like chessboards \cite{geiger2012automatic,zhou2018automatic,cui2020acsc}, spherical objects \cite{kummerle2018automatic} and polygonal boards \cite{park2014calibration}. However, their drawback is the offline applicability in controlled environments, limiting practicality in various real-world tasks. Recent efforts have aimed to overcome this limitation with more versatile targetless calibration methods, broadening usability across diverse real-world tasks by eliminating dependency on artificial auxiliary targets.

However, a prevalent assumption in existing LiDAR-camera calibration methods is that camera intrinsic parameters are constant during online robot operation. Yet, some practical tasks demand consideration for dynamic changes in intrinsic parameters. For instance, robots equipped with zoom lenses can dynamically adjust camera focal lengths to focus on specific regions of interest \cite{jovanvcevic2015automated}. Besides, camera focal length and principle point have a systematic relationship with temperature \cite{smith2010effects}, particularly affecting robots operating in varying weather conditions. In such cases, previous intrinsic parameters are no longer effective, necessitating a target-free approach to recover these altered parameters. However, accurate targetless calibration of both intrinsic and extrinsic parameters is still challenging: 1) In terms of intrinsic calibration, due to feature mismatches and the lack of prior knowledge of feature depths, target-free camera self-calibration methods are far less accurate than the chessboard-based ones. 2) Extrinsic calibration accuracy is compromised by inaccurate intrinsic parameters.

To address the above challenges, we propose a joint calibration method for LiDAR-camera systems in targetless environments. Our method builds upon a camera self-calibration approach \cite{schonberger2016structure} that recovers intrinsic parameters, camera poses and visual points through bundle adjustment. Furthermore, we introduce a new residual term based on the registration between visual points and LiDAR point cloud planes, forming a plane-constrained BA optimization. As the extrinsic parameters link visual points with LiDAR planes, they can be calibrated together with the intrinsic parameters via joint optimization. Besides, the planes help to filter out visual feature mismatches and provide depth constraints on visual points, which markedly enhances intrinsic calibration accuracy. In summary, our contributions are as follows:

\begin{itemize}
\item We propose an innovative and robust targetless method for joint intrinsic and extrinsic LiDAR-camera calibration. The joint calibration is formulated as a plane-constrained BA problem that concurrently optimizes intrinsic and extrinsic parameters.
\item We validate our approach through real-world experiments and comparisons with SOTA methods. Results demonstrate its robustness across calibration scenes and its superior accuracy over counterparts. Average rotation and translation errors are reduced to 0.12°/2.44 cm, with an average intrinsic error of 6.23 pixels for a 5-megapixel camera.
\item We make our codes publicly available on GitHub for the benefits of the community.
\end{itemize}

\vspace{-1mm}
\section{RELATED WORKS}

\subsection{Camera Self-Calibration}
\vspace{-1mm}

Camera self-calibration approaches aim to estimate intrinsic parameters using natural features rather than relying on known targets like checkerboards \cite{zhang2000flexible}. Such approaches (e.g. \cite{schonberger2016structure}, \cite{faugeras1992camera,maybank1992theory,luong1997self}) leverage sequences of images captured from diverse positions and orientations. By extracting and matching 2D features across images, multi-view geometry is employed to recover camera poses and intrinsic parameters. The prevalence of features in natural environments grants self-calibration approaches superior flexibility over target-based ones. However, the presence of feature mismatches and the absence of prior knowledge about feature depths make self-calibration approaches susceptible to observation noise and more prone to significant calibration errors.

\vspace{-1mm}
\subsection{Targetless LiDAR-Camera Extrinsic Calibration}
\vspace{-1mm}

Target-free LiDAR-camera calibration methods fall primarily into two categories. The first one exploits natural features in the environment, such as edges and data intensities. \cite{levinson2013automatic,kang2020automatic,yuan2021pixel,liu2022targetless} propose edge-alignment methods. In these works, natural 3D and 2D edges are extracted from LiDAR point clouds and camera images, respectively. Subsequently, the 3D edges are transformed into the camera frame and projected onto the image plane to compute cost functions. In \cite{pandey2015automatic}, the authors introduce a mutual information (MI) metric that characterizes the alignment between point cloud intensity and image intensity. By maximizing the MI metric, the extrinsic parameters are optimized. 

The second category is the motion-based method, often employing a coarse-to-fine calibration process. Works like \cite{taylor2016motion,ishikawa2018lidar,park2020spatiotemporal} initially recover extrinsic parameters through sensor ego-motion and subsequently refine them with appearance information. However, these approaches typically require extensive sensor movement to sufficiently activate the calibration.

\vspace{-1mm}
\subsection{Joint Intrinsic and Extrinsic LiDAR-Camera Calibration}
\vspace{-1mm}

Recent years have seen the emergence of methods tackling the joint LiDAR-camera calibration challenge. In \cite{kummerle2020unified}, a spherical auxiliary object is employed. The intrinsic and extrinsic parameters are simultaneously calibrated by minimizing distances from camera viewing rays and LiDAR points to the sphere surface. Similarly, \cite{yan2022joint} adopts a target-based strategy involving a plane target featuring a chessboard pattern and circular holes. The residual is the reprojection error of point cloud hole centers on the image plane.


For target-free solutions, \cite{miao2023coarse} designs an edge alignment method. They extract point cloud edges based on LiDAR intensity image, and match them with image edges. The intrinsic and extrinsic parameters are optimized by maximizing a probability density function. However, due to unstable LiDAR intensity measurements, the resulting intensity image is often noisy, leading to imprecise edge extraction. Another target-free approach \cite{tu2022multi} introduces a joint structure-from-motion (SfM) method. The overall cost function accounts for visual feature reprojection error, camera-to-LiDAR residual, and LiDAR-to-LiDAR residual, expressed as a weighted sum.

Our method is based on the plane-constrained BA optimization. As the normal vectors and depths of LiDAR planes can be precisely calculated, they can offer reliable depth constraints for visual points, thus enhancing intrinsic calibration accuracy. Besides, we incorporate the uncertainty models of visual points and features into the residuals, which further increases the calibration accuracy.

The concept of introducing LiDAR depth constraints on visual points in our work have been explored in the joint SfM-based method \cite{tu2022multi}. However, the important difference is that, \cite{tu2022multi} introduces edge features and LiDAR-to-LiDAR residuals. The extraction precision of curvature-based edges is much lower than that of plane, leading to inaccurate depth priors and point-to-edge registration. Moreover, the LiDAR-to-LiDAR residual in \cite{tu2022multi} is based on the pairwise registration manner, instead of the precise LiDAR bundle adjustment formulation, thus diminishing registration accuracy and ultimately the calibration accuracy. Notably, \cite{tu2022multi} neither shows higher intrinsic calibration accuracy than other methods, nor quantitatively reports the calibration precision. Therefore, the effectiveness of this approach remains unclear. In contrast, our work not only quantitatively achieves higher intrinsic calibration accuracy than counterparts, but also corroborates the effectiveness of LiDAR depth constraints on visual points.

\vspace{-1mm}
\section{METHODOLOGY}

\vspace{-1mm}
\subsection{Overview}
\vspace{-1mm}

\begin{figure}[t]
\centering
\includegraphics[width=0.85\linewidth]{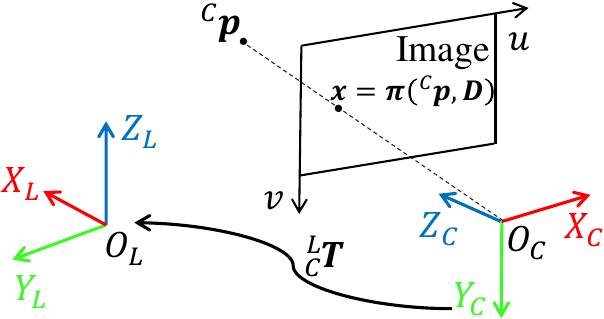}
\vspace{-4mm}
\caption{LiDAR and camera coordinate systems.}
\label{frame}
\vspace{-5mm}
\end{figure}

Fig. \ref{frame} illustrates the LiDAR and camera coordinate systems, along with the projection from a 3D point in the camera frame to a 2D pixel on the image plane. Specifically, we denote $L$ and $C$ as the LiDAR and camera frames, respectively. ${}^C \mathbf{p} \in \mathbb{R}^3$ is a 3D point expressed in the camera frame. ${}^L_C \mathbf T = ({}^L_C \mathbf R, {}^L_C \mathbf t) \in SE(3)$ denotes the extrinsic parameters from the camera frame to the LiDAR frame, while ${}^L \mathbf{p} = {}^L_C \mathbf{R} {}^C \mathbf{p} + {}^L_C \mathbf{t}$ projects the point into the LiDAR frame. $u$ and $v$ denote the abscissa axes within the image plane. Additionally, our method assumes the camera model to be the pinhole model with radical distortion. Concretely, $\mathbf{x} = \boldsymbol{\pi}({}^C \mathbf{p},\mathbf D)$ is the camera projection function that transforms ${}^C \mathbf{p}$ into a pixel $\mathbf{x} \in \mathbb{R}^2$ on the image plane. Here, $\mathbf{D}$ denotes the intrinsic parameters, including the focal length, principle point and radical distortion parameters. We further know that the function $\mathbf{f} = \boldsymbol{\pi}^{-1}(\mathbf{x},\mathbf{D})$ transforms an image pixel back into a 3D unit vector in the camera frame.

\begin{figure}[t]
\centering
\includegraphics[width=1.0\linewidth]{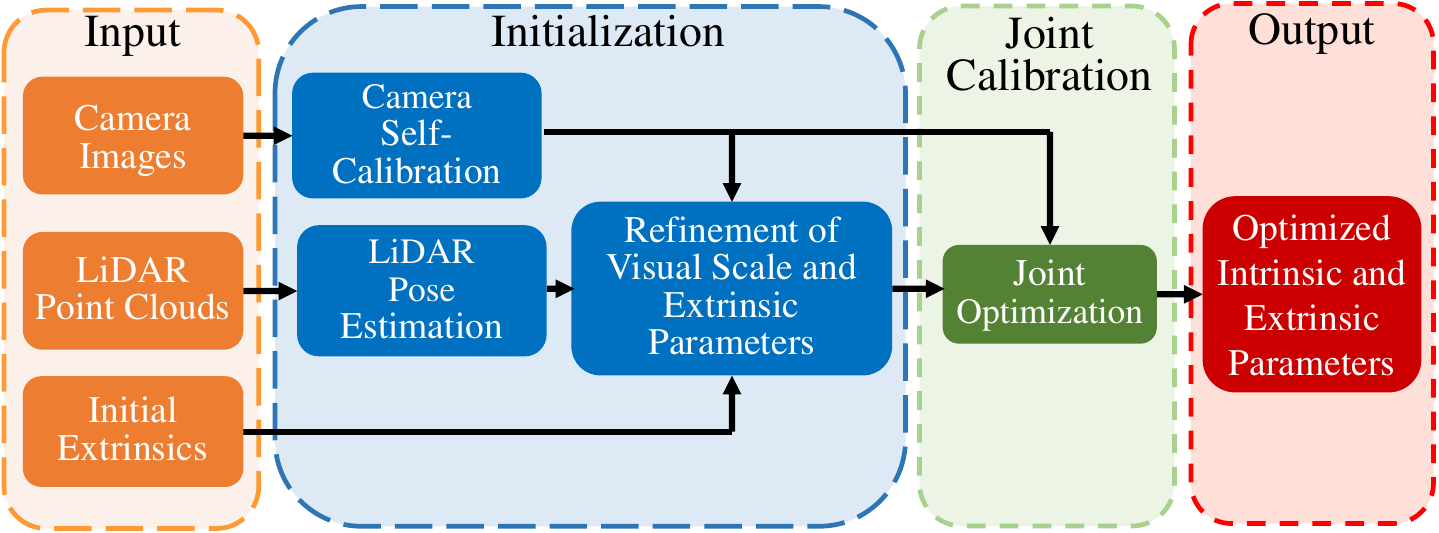}
\vspace{-7mm}
\caption{Calibration pipeline.}
\label{pipeline}
\vspace{-6mm}
\end{figure}

Our method conducts joint calibration of intrinsic parameters $\mathbf D$ and the extrinsic transformation ${}^L_C \mathbf T$. The process involves capturing multiple frames of camera images and LiDAR point clouds within the same scene. We denote $n$ as the number of frames, ${}^C \mathcal{T}=\{{}^C \mathbf{T}_1, \cdots, {}^C \mathbf{T}_n\}$ as the set of camera poses, and ${}^C \mathbf{T}_i=({}^C \mathbf{R}_i,{}^C \mathbf{t}_i) \in SE(3)$ as the $i$-th camera pose $(i=1,\cdots,n)$. Note that the camera world frame is the first camera frame, i.e. $({}^C \mathbf{R}_1,{}^C \mathbf{t}_1)=(\mathbf{I}_{3 \times 3},\mathbf{0}_{3 \times 1})$, where $\mathbf{I}$ is the identity matrix. Similarly, ${}^L \mathcal{T}=\{{}^L \mathbf{T}_1, \cdots, {}^L \mathbf{T}_n\}$ represents the set of LiDAR poses, ${}^L \mathbf{T}_i=({}^L \mathbf{R}_i,{}^L \mathbf{t}_i) \in SE(3)$ is the $i$-th LiDAR pose $(i=1,\cdots,n)$. The LiDAR world frame is the first LiDAR frame, and $({}^L \mathbf{R}_1,{}^L \mathbf{t}_1)=(\mathbf{I}_{3 \times 3},\mathbf{0}_{3 \times 1})$. In addition, the initial extrinsic parameters $({}^L_C \tilde{\mathbf{R}},{}^L_C \tilde{\mathbf{t}})$, which can be attained by hand-eye-calibration \cite{dornaika1998simultaneous}, are also required. The calibration pipeline takes raw sensor measurements and the initial extrinsic parameters as inputs, operating through a coarse-to-fine approach (see Fig. \ref{pipeline}). The initialization stage commences with camera self-calibration (Section \ref{chap3:sub2:subsec1}), which recovers intrinsic parameters, camera poses and visual points. Concurrently, LiDAR poses are estimated (Section. \ref{chap3:sub2:subsec2}). Afterwards, the visual scale and extrinsic parameters are iteratively refined (Section \ref{chap3:sub2:subsec3}) to provide good initial values for the subsequent stage. Finally, the joint calibration stage (see Section \ref{chap3:sub3}) optimizes the intrinsic and extrinsic parameters concurrently.

\vspace{-1mm}
\subsection{Initialization}
\vspace{-1mm}
\label{chap3:sub2}

\subsubsection{Camera Self-Calibration}
\label{chap3:sub2:subsec1}
Our approach obviates the need for pre-calibrating camera intrinsic parameters using target-based methods, instead coarsely recovering them through SfM. We employ the open-source software COLMAP \cite{schonberger2016structure} for this purpose. Initially, SIFT features \cite{lowe2004distinctive} are extracted and matched across images. The camera principle point is assumed to be the image center and initial distortion parameters are set to zeros. A range of focal length values is evaluated for absolute camera pose estimation. The value with the most 3D-2D correspondence inliers is chosen to initialize the focal length. Next, COLMAP incrementally registers all frames and refines parameters, including camera intrinsics, poses, and 3D points, via visual bundle adjustment. Upon registering all frames, we obtain camera poses ${}^C \mathcal{T}=\{{}^C \mathbf{T}_1, \cdots, {}^C \mathbf{T}_n\}$, visual points ${}^C \mathcal{P}=\{{}^C \mathbf{p}_1, \cdots, {}^C \mathbf{p}_m\}$, and camera intrinsic parameters $\mathbf D$. Besides, we calculate $\boldsymbol{\Sigma}_{{}^C \mathbf{p}_j}$, representing the covariance matrix (inverse of the Hessian matrix) of ${}^C \mathbf{p}_j$, after BA. Notably, both camera poses and visual points are expressed in the first camera frame. Additionally, feature $\mathbf{x}_{ij}$ is the observation of ${}^C \mathbf{p}_j$ in the $i$-th image.

\subsubsection{LiDAR Pose Estimation}
\label{chap3:sub2:subsec2}
Accurate estimation of LiDAR poses ${}^L \mathcal{T}$ is pivotal for subsequent pairwise registration between visual points and LiDAR point clouds (see Section \ref{chap3:sub2:subsec3}). Our approach commences by incrementally estimating each frame's pose using point-to-plane ICP \cite{low2004linear}. Building upon this, LiDAR bundle adjustment \cite{liu2022efficient} is executed to optimize all LiDAR poses. This process minimizes feature distances to the planes, thereby mitigating pose drift and enhancing overall accuracy.

\subsubsection{Refinement of Visual Scale and Extrinsic Parameters}
\label{chap3:sub2:subsec3}
The scale of the monocular SfM is ambiguous and 
needs to be recovered to the real-world scale. In addition, the initial extrinsic parameters $({}^L_C \tilde{\mathbf{R}},{}^L_C \tilde{\mathbf{t}})$ need to be refined to ensure correct feature matching in the joint calibration stage. 

According to \cite{park2020spatiotemporal}, we derive the coarse visual scale $\tilde{s}$ by solving the linear equation:
\begin{equation}
\setlength{\abovedisplayskip}{2pt}
\setlength{\belowdisplayskip}{2pt}
    \tilde{s} {}^L_C \tilde{\mathbf{R}} {}^C \mathbf{t}_i
    =
    {}^L \mathbf{t}_i - (\mathbf{I}-{}^L \mathbf{R}_i)
    {}^L_C \tilde{\mathbf{t}},
    \label{eq_scale}
\end{equation}
where $\mathbf I \in \mathbb{R}^{3 \times 3}$ is the identity matrix. Later, we iteratively refine both the visual scale and extrinsic parameters using point-to-plane registration. We denote $\kappa\;(\kappa=1,\cdots,\kappa_{max})$ the iteration index, $\hat{s}^{\kappa}$ the scale, and $({}^L_C \hat{\mathbf{R}}^{\kappa},{}^L_C \hat{\mathbf{t}}^{\kappa})$ the extrinsic parameters in the $\kappa$-th iteration. Besides, we denote ${}^C \hat{\mathbf{p}}_{j}^{\kappa}\;(j=1,\cdots,m)$ the visual point in the $\kappa$-th iteration, computed as: 
\begin{equation}
\setlength{\abovedisplayskip}{2pt}
\setlength{\belowdisplayskip}{2pt}
    {}^C \hat{\mathbf{p}}_{j}^{\kappa}
    =
    \hat{s}^{\kappa}
    {}^C \mathbf{p}_{j}.
    \label{eq_update_point}
\end{equation}
Creating point-to-plane correspondences involves transforming visual points into each LiDAR frame:
\begin{equation}
\setlength{\abovedisplayskip}{2pt}
\setlength{\belowdisplayskip}{2pt}
    {}^{L_i} \hat{\mathbf{p}}_{j}^{\kappa}
    =
    {}^L \mathbf{R}_i
    ({}^L_C \hat{\mathbf{R}}^{\kappa}
    {}^C \hat{\mathbf{p}}_{j}^{\kappa}
    +
    {}^L_C \hat{\mathbf{t}}^{\kappa})
    +
    {}^L \mathbf{t}_i,
    \label{eq_transform_to_lidar}
\end{equation}
and searching for its $l$ nearest neighbours $\mathbf Q_{ij}^{\kappa}=\{\mathbf{q}_{ijk}^{\kappa}; k = 1, \cdots, l\}$ in the LiDAR point cloud. The covariance matrix of the point cloud is constructed as:
\begin{equation}
\setlength{\abovedisplayskip}{2pt}
\setlength{\belowdisplayskip}{2pt}    \mathbf{A}_{ij}^{\kappa} 
    = 
    \frac{1}{l}
    \sum_{k=1}^{l} 
    (\mathbf{q}_{ijk}^{\kappa}
    - 
    \bar{\mathbf{q}}_{ij}^{\kappa}) 
    (\mathbf{q}_{ijk}^{\kappa} 
    - 
    \bar{\mathbf{q}}_{ij}^{\kappa})^T,
    \label{eq_cov_matrix}
\end{equation}
where $\bar{\mathbf{q}}_{ij}^{\kappa}=\frac{1}{l}\sum_{k=1}^{l} \mathbf{q}_{ijk}^{\kappa}$ is the point cloud center. We denote ${\lambda}_1$, ${\lambda}_2$, ${\lambda}_3$ the eigenvalues of $\mathbf{A}_{ij}^{\kappa}$, and $\lambda_1 \geq \lambda_2 \geq \lambda_3$. If ${\lambda}_2/{\lambda}_3$ exceeds a given threshold, a valid point-to-plane constraint is established. Denoting $\mathcal{V}^{\kappa}$ as the set of all valid point-to-plane pairs, optimization is defined as:
\begin{equation}
\setlength{\abovedisplayskip}{2pt}
\setlength{\belowdisplayskip}{2pt}    \mathop{\min}_{\hat{s}^{\kappa},
    {}^L_C \hat{\mathbf{R}}^{\kappa},
    {}^L_C \hat{\mathbf{t}}^{\kappa}}
    \sum_{(i,j) \in \mathcal{V}^{\kappa}}
    H({\mathbf{n}_{ij}^{\kappa}}^{T}
    ({}^{L_i} \hat{\mathbf{p}}_{j}^{\kappa}
    -\bar{\mathbf{q}}_{ij}^{\kappa})),
    \label{eq_refine_opt}
\end{equation}
where $H(\cdot)$ is the Huber loss function \cite{huber1992robust} and $\mathbf{n}_{ij}^{\kappa}$ indicates the plane normal vector, i.e.\ the eigenvector corresponding to ${\lambda}_3$. The optimization is solved by the Ceres Solver\footnote[3]{\url{http://ceres-solver.org}}. 

The iterative refinement initiates with the initial values: $\hat{s}^{1} = \tilde{s}$ and $ 
({}^L_C \hat{\mathbf{R}}^{1},{}^L_C \hat{\mathbf{t}}^{1})=
({}^L_C \tilde{\mathbf{R}},{}^L_C \tilde{\mathbf{t}})$. 
Optimized parameters $(\hat{s}^{\kappa})^{*}, {{}^L_C \hat{\mathbf{R}}^{\kappa}}^{*}, {{}^L_C \hat{\mathbf{t}}^{\kappa}}^{*}$ in the $\kappa$-th iteration are used as initial values for the next iteration. After the last iteration, we update camera poses, visual points, point covariance matrices using the final optimized scale $(\hat{s}^{\kappa_{max}})^{*}$.

\vspace{-2mm}
\subsection{Joint Calibration}
\vspace{-1mm}
\label{chap3:sub3}



As mentioned in Section \ref{chap3:sub2:subsec1}, the intrinsic parameters are recovered by visual bundle adjustment. However, mainly due to the lack of prior knowledge of visual point depths, the intrinsic parameters are prone to large calibration error. To mitigate this problem, we introduce the LiDAR planes to offer accurate depth priors. This is achieved by adding another residual term into the BA: the registration residual between visual points and LiDAR planes. This term effectively reduces the point-to-plane distances, aiding in reducing the intrinsic calibration error. Besides, as the extrinsic parameters are also constrained by the point-to-plane registration, they can be optimized together with intrinsic parameters, to minimize the integrated cost function.

We incorporate the uncertainty models of visual points and features to derive the residual terms as follows.

\subsubsection{Point-to-plane residuals}
\label{chap3:sub3:subsec1}
For each point ${}^C \mathbf{p}_j$, we project it to every LiDAR frame and search for its nearest-neighbour point cloud. We use the same criteria mentioned in Section \ref{chap3:sub2:subsec3} to determine whether ${}^C \mathbf{p}_j$ is actually associated with a plane. Besides, as a large point-to-plane distance usually indicates image feature mismatches, the point-to-plane correspondence is discarded if the distance exceeds a predefined threshold. Suppose ${}^C \mathbf{p}_j$ finds a valid corresponding plane $(\mathbf{n}_{ij}, \bar{\mathbf{q}}_{ij})$ in the $i$-th LiDAR frame, the residual is derived as follows. We denote ${}^{L_i}_C \mathbf{T} = {}^L_C \mathbf{T} {}^C \mathbf{T}_i = ({}^{L_i}_C \mathbf{R},{}^{L_i}_C \mathbf{t})$ the $SE(3)$ transformation from the first camera frame to the $i$-th LiDAR frame. Because ${}^C \mathbf{p}_j$ is subject to Gaussian noise: ${}^C \mathbf{p}_j
={}^C \mathbf{p}_j^{gt}+
\boldsymbol{\delta}_{{}^C \mathbf{p}_j}$, where ${}^C \mathbf{p}_j^{gt}$ is the ground-true point location and $\boldsymbol{\delta}_{{}^C \mathbf{p}_j} \sim \mathcal{N}(\mathbf{0}_{3 \times 1}, \boldsymbol{\Sigma}_{{}^C \mathbf{p}_j})$ is the point uncertainty, the projection of ${}^C \mathbf{p}_j$ in the $i$-th LiDAR frame can expressed as
\begin{equation}
\setlength{\abovedisplayskip}{2pt}
\setlength{\belowdisplayskip}{2pt}
    {}^{L_i} \mathbf{p}_j
    =
    {}^{L_i}_C \mathbf{R}
    {}^C \mathbf{p}_j +
    {}^{L_i}_C \mathbf{t}
    =
    {}^{L_i} \mathbf{p}_j^{gt} +
    \boldsymbol{\delta}_{{}^{L_i} \mathbf{p}_j},
    \label{eq_point_noise_lidar}
\end{equation}
where ${}^{L_i} \mathbf{p}_j^{gt} = {}^{L_i}_C \mathbf{R} {}^C \mathbf{p}_j^{gt} + {}^{L_i}_C \mathbf{t}$, $\boldsymbol{\delta}_{{}^{L_i} \mathbf{p}_j} \sim \mathcal{N}(\mathbf{0}_{3 \times 1}, \boldsymbol{\Sigma}_{{}^{L_i} \mathbf{p}_j})$ and $\boldsymbol{\Sigma}_{{}^{L_i} \mathbf{p}_j} = {}^{L_i}_C \mathbf{R} \boldsymbol{\Sigma}_{{}^C \mathbf{p}_j} {}^{L_i}_C \mathbf{R}^T$. Since the ground-true point ${}^{L_i} \mathbf{p}_j^{gt}$ lies exactly on the LiDAR plane $(\mathbf{n}_{ij}, \bar{\mathbf{q}}_{ij})$, we have
\begin{equation}
\setlength{\abovedisplayskip}{2pt}
\setlength{\belowdisplayskip}{2pt}
    \mathbf{n}_{ij}^T
    ({}^{L_i} \mathbf{p}_j^{gt}-\bar{\mathbf{q}}_{ij})
    =
    0.
    \label{eq_gt_point}
\end{equation}
By combining (\ref{eq_point_noise_lidar}) and (\ref{eq_gt_point}), the actual point-to-plane distance can be expressed by
\begin{equation}
\setlength{\abovedisplayskip}{2pt}
\setlength{\belowdisplayskip}{2pt}
    \mathbf{n}_{ij}^T
    ({}^{L_i} \mathbf{p}_j
    - \bar{\mathbf{q}}_{ij})
    =
    \mathbf{n}_{ij}^T
    \boldsymbol{\delta}_{{}^{L_i} \mathbf{p}_j}
    \sim
    \mathcal{N}(0, 
    \mathbf{n}_{ij}^T
    \boldsymbol{\Sigma}_{{}^{L_i} \mathbf{p}_j}
    \mathbf{n}_{ij}).
    \label{eq_distance}
\end{equation}
Applying maximum likelihood estimation (MLE), the overall point-to-plane residual $E^{P}$ is formulated as:
\begin{equation}
\setlength{\abovedisplayskip}{2pt}
\setlength{\belowdisplayskip}{2pt}
    E^{P} 
    =
    \frac{1}{2}
    \sum_{(i,j) \in \mathcal{V}^{P}}
    \frac
    {(\mathbf{n}_{ij}^T
     ({}^{L_i} \mathbf{p}_j
    - \bar{\mathbf{q}}_{ij}))^2}
    {\mathbf{n}_{ij}^T
    \boldsymbol{\Sigma}_{{}^{L_i} \mathbf{p}_j}
    \mathbf{n}_{ij}},
    \label{eq_distance_residuals}
\end{equation}
where $\mathcal{V}^{P}$ denotes the set of valid point-to-plane pairs. 

\subsubsection{Visual reprojection residuals}
As we propose to use visual points associated with LiDAR planes, points without any point-to-plane correspondences are discarded. Suppose point ${}^C \mathbf{p}_j$ successfully finds at least one plane correspondence. 
Visual feature $\mathbf{x}_{i,j}$ indicates that ${}^C \mathbf{p}_j$ is observed by the $i$-th image. Denote ${}^{C_{i}} \mathbf{p}_j = {}^C \mathbf{R}_i {}^C \mathbf{p}_j + {}^C \mathbf{t}_i$ the projection of ${}^C \mathbf{p}_j$ in the $i$-th image frame, the actual projection of ${}^C \mathbf{p}_j$ on the image is $\hat{\mathbf{x}}_{ij}=\boldsymbol{\pi}({}^{C_{i}} \mathbf{p}_j,\mathbf{D})$. In practice, the measurement is normally subject to a Gaussian noise, i.e. $\mathbf{x}_{ij}=   \mathbf{x}_{ij}^{gt}+
\boldsymbol{\delta}_{\mathbf{x}_{ij}}$,
where $\boldsymbol{\delta}_{\mathbf{x}_{ij}} \sim  \mathcal{N}(\mathbf{0}_{2 \times 1},\boldsymbol{\Sigma}_{\mathbf{x}_{ij}})$. Again, through MLE, the overall visual reprojection residual $E^{V}$ is formulated as:
\begin{equation}
\setlength{\abovedisplayskip}{3pt}
\setlength{\belowdisplayskip}{2pt}
    E^{V} 
    =
    \frac{1}{2}
    \sum_{(i,j) \in \mathcal{V}^{V}}
    (\mathbf{x}_{ij}-\hat{\mathbf{x}}_{ij})^T \boldsymbol{\Sigma}_{\mathbf{x}_{ij}}^{-1}(\mathbf{x}_{ij}-\hat{\mathbf{x}}_{ij}),
    \label{eq_reprojection_residuals}
\end{equation}
where $\mathcal{V}^{V}$ denotes the set of valid feature observation pairs.

Finally, a weighting factor $\alpha$ is introduced to further balance the two residual terms. The optimal calibration parameters, camera poses and visual points are obtained by minimizing the weighted cost function:
\begin{equation}
\setlength{\abovedisplayskip}{2pt}
\setlength{\belowdisplayskip}{2pt}
    {}^L_C \mathbf{T}^*,\mathbf{D}^*,
    {}^C \mathcal{T}^*,{}^C \mathcal{P}^*
    =
    \mathop{\arg}\mathop{\min}_{{}^L_C \mathbf{T},\mathbf{D},
    {}^C \mathcal{T},{}^C \mathcal{P}}
    (\alpha E^P
    +
    E^V),
    \label{eq_final_opt}
\end{equation}
which is solved by the Levenberg-Marquardt algorithm \cite{more2006levenberg} with the Ceres Solver.


\vspace{-2mm}
\section{EXPERIMENTS}
\vspace{-1mm}
In this section, we validate our calibration method and compare it against other state-of-the-art approaches on two real-world datasets: the HKU campus dataset and the KITTI dataset \cite{geiger2012we}.

For quantitative performance comparison, we calculate the calibration error between the calibrated parameters and the ground-true values. Note that the ground-true intrinsic and extrinsic parameters are both obtained by standard target-based methods. We denote ${}^L_C \mathbf{R}^{gt}$, ${}^L_C \mathbf{t}^{gt}$, $\mathbf{D}^{gt}$ the ground-true rotation, translation and camera intrinsic parameters, respectively. The rotation error is given by $\mathcal{E}_\mathbf{R}=\arccos{(\frac{1}{2}tr({}^L_C \mathbf{R}^*({}^L_C \mathbf{R}^{gt})^T) - \frac{1}{2})}$, expressed in degrees. The translation error is given by $\mathcal{E}_\mathbf{t} = \left\|{}^L_C \mathbf{t}^* - {}^L_C \mathbf{t}^{gt} \right\|$, expressed in centimeters. For intrinsic parameters, their coupling effect on determining pixel locations makes direct parameter-wise comparison less meaningful. Instead, we use the average reprojection error of image pixels. With the image width and height denoted by $w$ and $h$ pixels respectively, the intrinsic calibration error, expressed in pixels, is given by 
\begin{equation}
\setlength{\abovedisplayskip}{2pt}
\setlength{\thinmuskip}{1mu}
\setlength{\medmuskip}{1mu}
\setlength{\thickmuskip}{1mu}
\hspace{-2mm}
    \mathcal{E}_\mathbf{D}
    =
    \frac{1}{wh} \sum_{u=1}^{w} \sum_{v=1}^{h}
    \left\|\boldsymbol{\pi}(\boldsymbol{\pi}^{-1}([u,v]^T,\mathbf{D}^{gt}),\mathbf{D}^*)-[u,v]^T\right\|.
    \label{intrinsic_error}
\end{equation}

\subsection{HKU Campus Dataset}
\vspace{-1mm}
\label{chap4:sub1}

\begin{figure}[t]
\centering
\includegraphics[width=0.90\linewidth]{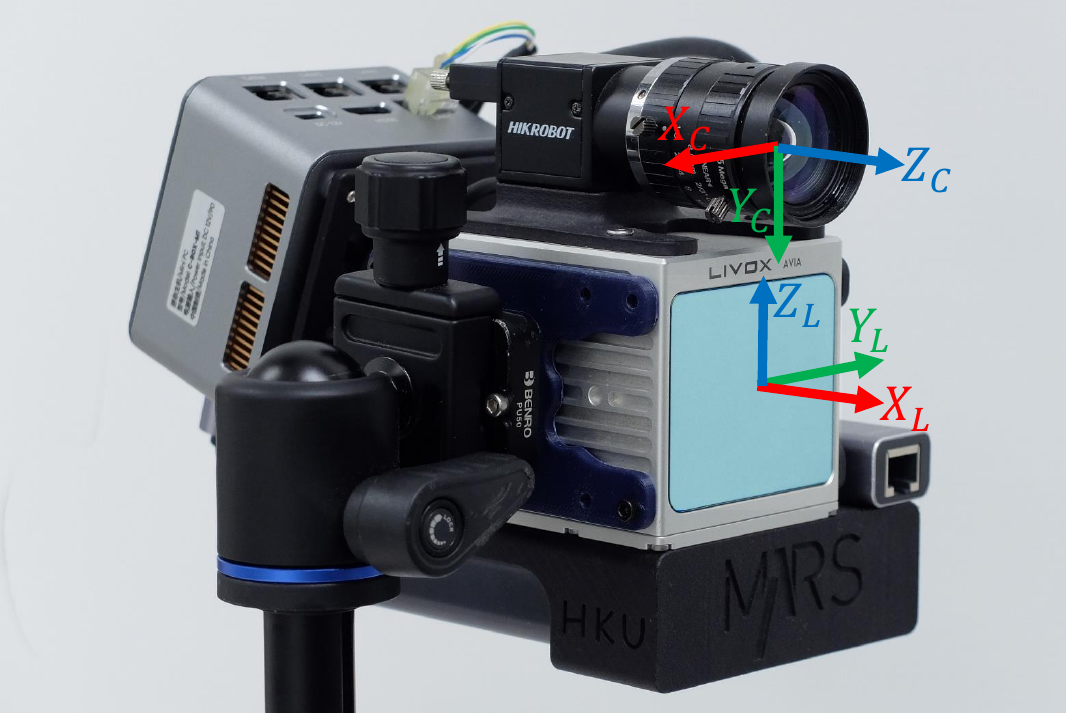}
\vspace{-3mm}
\caption{LiDAR-camera sensor suite. The LiDAR type and camera type are LIVOX AVIA and HIKVISION MV-CA050-11UC, respectively.}
\label{sensorsuite}
\vspace{-1mm}
\end{figure}

This dataset is collected in The University of Hong Kong campus, hence the name. The device for data collection is shown in Fig. \ref{sensorsuite}. The sensor suite comprises of a high-resolution solid-state LiDAR named LIVOX AVIA\footnote[4]{\url{https://www.livoxtech.com/avia}}, and an industry camera HIKVISION MV-CA050-11UC\footnote[5]{\url{https://www.hikvisionweb.com/product/camera/usb3-0/mv-ca050-11uc/}} with a resolution of 5 megapixels.

\begin{figure}[t]
\centering
\includegraphics[width=0.95\linewidth]{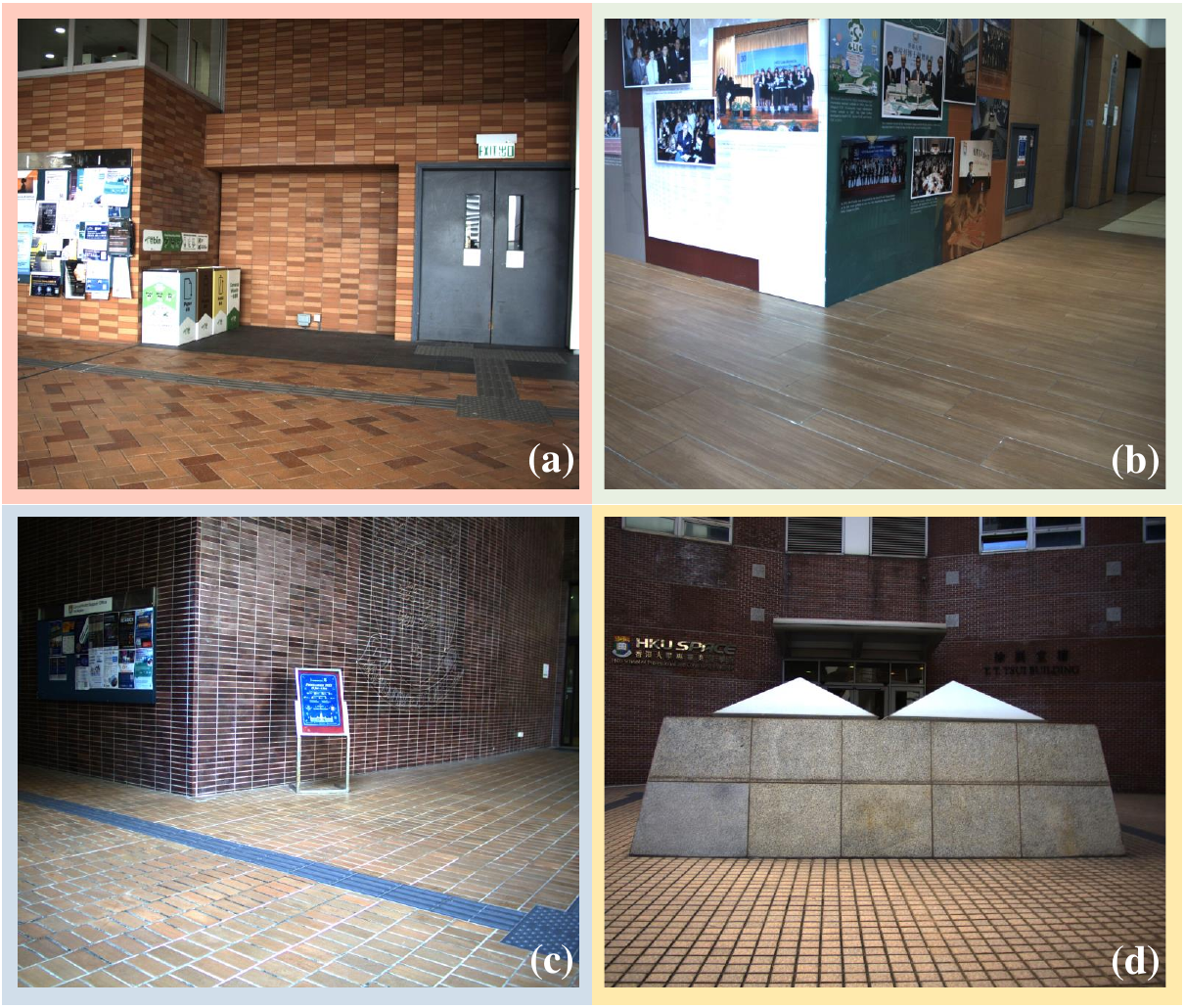}
\vspace{-3mm}
\caption{Calibration scenes in HKU campus dataset. (a)-(d) Scene 1-4.}
\label{scene}
\vspace{-6mm}
\end{figure}

The data are collected from 4 different scenes (see Fig. \ref{scene}). For each scene, we capture 8 data frames from different positions and orientations. To avoid point cloud distortion and sensor synchronization issues, we keep the sensor suite static when recording each data frame. In addition, to make full use of the non-repetitive scanning characteristic \cite{liu2021low} of high-resolution LiDARs, each point cloud is accumulated for 3 seconds to get a dense representation of the surroundings.

Before running the calibration algorithm, we first attain the initial rotation ${}^L_C \tilde{\mathbf R}$ and translation ${}^L_C \tilde{\mathbf t}$. We move the sensor suite along a sufficiently excited continuous trajectory, and estimate the LiDAR and camera odometry by \cite{xu2022fast} and \cite{schonberger2016structure}, respectively. Then, $({}^L_C \tilde{\mathbf R},{}^L_C \tilde{\mathbf t})$ are solved by hand-eye-calibration \cite{dornaika1998simultaneous}. These initial extrinsic parameters are then applied across all calibration scenes. The calculated initial rotation and translation errors are 4.81$^\circ$/43.94 cm. Besides, we resort to the MATLAB Camera Calibration toolbox\footnote[6]{\url{https://ww2.mathworks.cn/help/vision/camera-calibration.html}} to acquire the ground-true intrinsic parameters $\mathbf{D}^{gt}$, and adopts a target-based approach \cite{zhou2018automatic} to obtain ${}^L_C \mathbf{R}^{gt}$, ${}^L_C \mathbf{t}^{gt}$.

We compare our method with several open-source target-free calibration approaches. In particular, we compare with a depth edge-based approach \cite{yuan2021pixel}, which uses the depth-continuous edges in the point cloud, on extrinsic calibration; we compare with an intensity edge-based co-calibration method \cite{miao2023coarse}, which extracts point cloud edges based on LiDAR intensity image, on both intrinsic and extrinsic calibration. Besides, as the SfM-based method \cite{schonberger2016structure} is used in our initialization stage to initialize intrinsic parameters (see Section \ref{chap3:sub2:subsec1}), we compare with it on intrinsic calibration. \cite{yuan2021pixel,miao2023coarse} and our method share the same initial extrinsic parameters $({}^L_C \tilde{\mathbf R}, {}^L_C \tilde{\mathbf t})$. In addition, since \cite{yuan2021pixel} does not optimize intrinsic parameters, its intrinsic parameters are kept as the ground-true values $\mathbf{D}^{gt}$; the initial intrinsic parameters of \cite{miao2023coarse} are obtained by \cite{schonberger2016structure}. The parameters of each method are tuned to obtain the best results to the authors' efforts. Each method utilizes the same parameters across all the scenes.

\begin{table}[h]
    \centering
    \caption{Intrinsic calibration errors (pixels) on HKU campus dataset}
    \vspace{-4mm}
    \label{tab:hku_campus_in}
    \setlength\tabcolsep{2.0pt}
    \begin{tabular}{cccccc}
    \toprule
    & Scene 1 & Scene 2 & Scene 3 & Scene 4 & Average \\
    \midrule
    \textbf{Ours} & \textbf{5.58} & \textbf{3.83} & \textbf{7.57} & \textbf{7.93} & \textbf{6.23} \\ 
    SfM-Based \cite{schonberger2016structure} & 9.12 & 9.18 & 12.26 & 14.27 & 11.21 \\
    Intensity Edge-Based \cite{miao2023coarse} & 23.80 & 14.84 & 25.00 & 32.24 & 23.97 
 \\
    \bottomrule
    \end{tabular}
\vspace{-5mm}
\end{table}

\begin{table}[h]
    \centering
    \caption{Rotation errors (degrees) and translation errors (centimeters) on HKU campus dataset}
    \vspace{-3mm}
    \label{tab:hku_campus_ex}
    \setlength\tabcolsep{2.5pt}
    \begin{tabular}{cccccc}
    \toprule
    & Scene 1 & Scene 2 & Scene 3 & Scene 4 & Average \\
    \midrule
    \textbf{Ours} & \textbf{0.08}/\textbf{1.94} & \textbf{0.09}/1.91 & \textbf{0.16}/\textbf{2.75} & \textbf{0.17}/\textbf{3.16} & \textbf{0.12}/\textbf{2.44} \\ 
    \makecell{Depth Edge-\\Based \cite{yuan2021pixel}} & 0.48/6.84 & 1.93/9.59 & 0.31/8.14 & 2.59/62.99 & 1.33/21.89 \\
    \makecell{Intensity Edge-\\Based \cite{miao2023coarse}} & 0.44/5.86 & 0.53/\textbf{0.95} & 1.96/12.74 & 5.04/57.82 & 2.00/19.34 \\
    \bottomrule
    \end{tabular}
    \begin{tablenotes}
        \footnotesize
        \item $^*$The initial rotation and translation errors are 4.81$^{\circ}$/43.94 cm.
    \end{tablenotes}    
\vspace{-2mm}
\end{table}

The experimental results are summarized in Table \ref{tab:hku_campus_in} (intrinsic calibration) and Table \ref{tab:hku_campus_ex} (extrinsic calibration). Regarding intrinsic calibration, our method achieves the highest accuracy across all scenes. Most notably, our average intrinsic error (6.23 pixels) significantly outperforms the SfM-based method \cite{schonberger2016structure} (11.21 pixels). Because the intrinsic parameters calibrated by \cite{schonberger2016structure} serve as the initial values for our joint optimization step (see Section \ref{chap3:sub2:subsec3}), this comparison significantly verifies that enforcing depth constraints from LiDAR point cloud planes to visual points does enhance the intrinsic calibration accuracy. As for extrinsic calibration, our method shows accurate results across the 4 scenes. In contrast, due to numerous feature mismatches stemming from initial extrinsic parameters, the depth edge-based method \cite{yuan2021pixel} fails on scene 2 and 4, while the intensity edge-based method \cite{miao2023coarse} fails on scene 3 and 4. This underscores that our method is more robust to the variety of calibration scenes. Moreover, Even in scenes where \cite{yuan2021pixel,miao2023coarse} successfully calibrate extrinsic parameters, our method still achieves superior accuracy, except for a slightly larger translation error in scene 2 compared to \cite{miao2023coarse}. There are two reasons for our superior performance. Firstly, the extraction of point cloud planes is more precise than that of edges \cite{miao2023coarse,yuan2021pixel},  introducing minimal noise during optimization. Secondly, our method effectively eliminates outliers based on point cloud eigenvalues and point-to-plane distances (see Section \ref{chap3:sub3:subsec1}). In contrast, \cite{miao2023coarse,yuan2021pixel} lack effective strategies to filter out edge mismatches, allowing these mismatches to influence the optimization process, thus degrading calibration accuracy.

\begin{figure*}[t]
\centering
\includegraphics[width=0.85\linewidth]{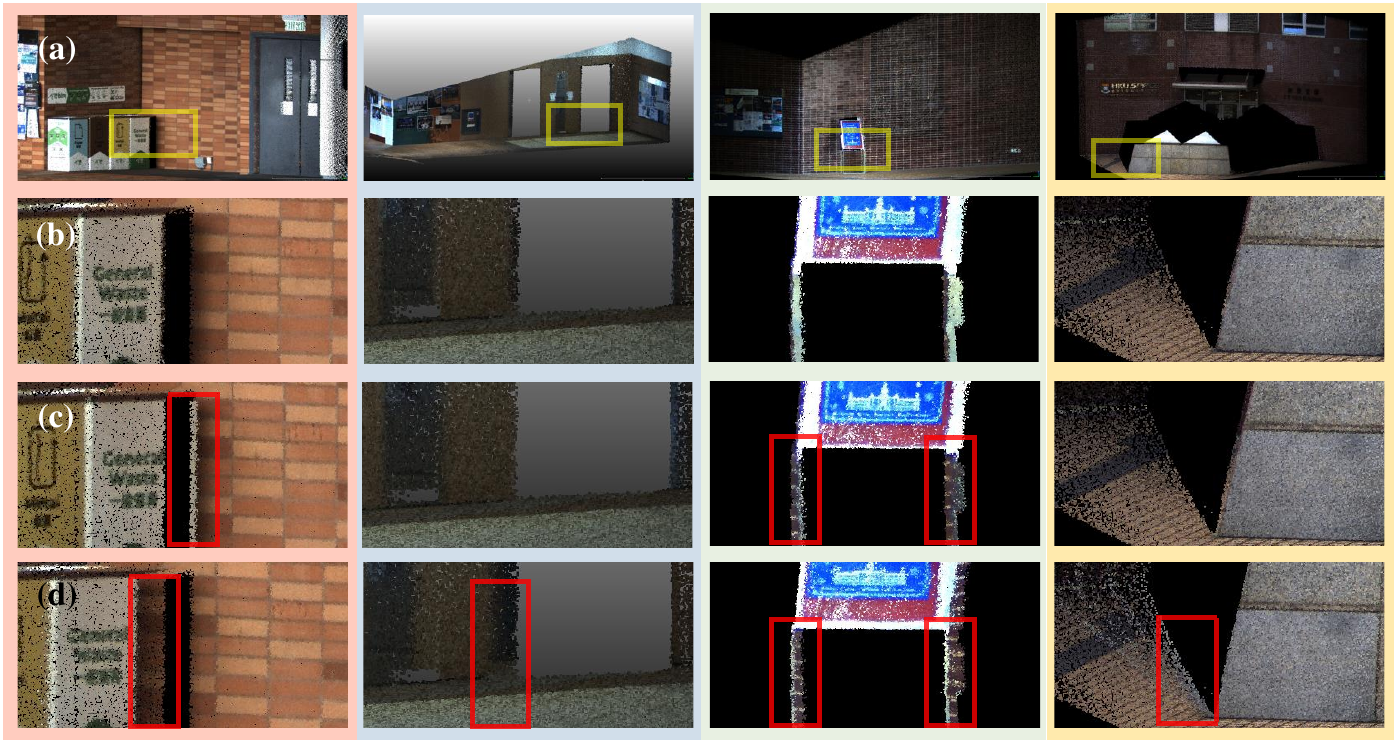}
\vspace{-3mm}
\caption{(a) Colorized point clouds using ground-true calibration parameters. Regions in the yellow boxes are used for comparison. (b) Colorization by our calibration result. (c) Colorization by \cite{miao2023coarse}'s calibration result. (d) Colorization by \cite{yuan2021pixel}'s calibration result. Red boxes indicate color misplacement.}
\label{colorization}
\vspace{-1mm}
\end{figure*}

In addition to the quantitative comparison, we further assess the calibration quality by colorizing LiDAR point clouds using the calibration results (see Fig. \ref{colorization}). To ensure a fair comparison, we employ the best calibration results from each method across the four scenes for colorization. Specifically, for our method and the intensity edge-based one \cite{miao2023coarse}, we utilize the parameters from scene 2, given their lowest intrinsic and translation errors; for the depth edge-based method \cite{yuan2021pixel}, we employ the parameters from scene 3 due to their minimal rotation error. Fig. \ref{colorization} reveals that our method consistently achieves high-quality colorization results across all scenes, while \cite{miao2023coarse} and \cite{yuan2021pixel} show discernible color misplacement for certain scenes. This visual assessment further substantiates that our method surpasses other SOTA approaches in terms of accuracy and reliability.

\begin{table*}[t]
    \centering
    \caption{Rotation errors (degrees) and translation errors (centimeters) on KITTI dataset}
    \vspace{-3mm}
    \label{tab:kitti}
    \setlength\tabcolsep{3pt}
    \begin{tabular*}{\linewidth}{ccccccccccccc}
    \toprule
     & 00 & 01$^*$ & 02 & 03 & 04$^*$ & 05 & 06 & 07 & 08 & 09 & 10 & Average \\
    \midrule
    Ours & \textbf{0.13}/\textbf{2.28} & \textbf{0.21}/17.05 & \textbf{0.16}/8.98 & \textbf{0.16}/4.85 & \textbf{0.14}/39.45 & 0.14/\textbf{1.12} & \textbf{0.14}/3.68 & \textbf{0.11}/\textbf{1.73} & \textbf{0.07}/\textbf{2.48} & \textbf{0.07}/5.60 & \textbf{0.16}/\textbf{3.77} & \textbf{0.13}/3.83 \\
    Joint SfM \cite{tu2022multi} & 0.16/4.40 & 0.22/\textbf{6.20} & 0.22/\textbf{5.10} & 0.59/\textbf{2.11} & 0.45/\textbf{2.76} & \textbf{0.13}/1.81 & 0.29/\textbf{2.20} & 0.56/2.35 & 0.19/4.33 & 0.75/\textbf{3.73} & 0.39/4.41 & 0.37/\textbf{3.38} \\
    \bottomrule
    \end{tabular*}
    \begin{tablenotes}
        \footnotesize
        \item $^*$Calibration results on sequence 01 and 04 are not used to calculate the average errors.
    \end{tablenotes}
\vspace{-6mm}
\end{table*}

\vspace{-1mm}
\subsection{KITTI Dataset}
\vspace{-1mm}

To assess the efficacy of our method's extrinsic calibration in diverse and cluttered environments, we conduct experiments on the KITTI odometry dataset \cite{geiger2012we}. The dataset comprises 11 sequences collected on urban driveway. The LiDAR (Velodyne HDL-64E) is based on the traditional mechanical scanning pattern, with 64 laser beams. Notably, the dataset provides ${}^L_C \mathbf{R}^{gt}$, ${}^L_C \mathbf{t}^{gt}$ and $\mathbf{D}^{gt}$.

We evaluate our method across all 11 sequences, and benchmark its performance against the joint SfM approach \cite{tu2022multi} which combines the camera-to-camera, camera-to-LiDAR and LiDAR-to-LiDAR residuals together. For a fair comparison, we initialize the extrinsic parameters by randomly perturbing the ground-true values (2$^{\circ}$ for ${}^L_C \mathbf{R}^{gt}$; 20 cm for ${}^L_C \mathbf{t}^{gt}$), in the same way as \cite{tu2022multi} does. Besides, the intrinsic parameters remain consistent with the ground-truth values for both methods. The results are shown in Table \ref{tab:kitti}, where we directly list the calibration errors reported in \cite{tu2022multi}. We notice that in sequence 01 and 04, the registration between visual points and LiDAR planes encounters a lack of constraints along the $Z$ axis in the camera frame. This is a typical degenerate case in point-to-plane registration (see \cite{ramalingam2010p}), rendering these sequences unsuitable for our method's calibration criteria. Therefore, the calibration results from these sequences are excluded from the calculation of average errors. As shown in the table, although our average translation error (3.83 cm) is marginally higher than that of \cite{tu2022multi} (3.38 cm), our method excels in rotation calibration. Specifically, we achieve superior rotation accuracy across almost all sequences except sequence 05. The overall rotation error of 0.13$^\circ$ significantly surpasses the performance of \cite{tu2022multi} (0.37$^\circ$), signifying a remarkable reduction of approximately two-thirds. When applying the $SE(3)$ transformation on a 3D point, a rotation difference of 0.24$^\circ$ typically results in considerably larger misplacement compared to a translation difference of 0.45 cm. Overall, our method achieves better calibration accuracy than our counterpart, reaffirming its adaptability to challenging scenes.



\section{CONCLUSIONS}
\vspace{-1mm}

In this paper, we propose a novel target-free approach for the joint calibration of LiDAR-camera systems. The core innovation lies in our formulation of joint calibration as a plane-constrained bundle adjustment problem, enriched by the integration of uncertainty models for visual features and triangulated points. Our method's efficacy is substantiated through a series of real-world experiments, demonstrating its superiority in accuracy and robustness compared to SOTA methods. Notably, our approach maintains exceptional calibration accuracy even in highly demanding environments.





\clearpage
\bibliographystyle{IEEEtran}
\bibliography{reference}

\end{document}